%% file: main.tex
\documentclass{article}

\usepackage{graphicx}
\usepackage{xspace}
\usepackage{tabularx}
\usepackage{multirow}
\usepackage{makecell}
\usepackage{amsmath} 
\usepackage{algorithm}
\usepackage{algorithmicx}
\usepackage{algpseudocode}
\usepackage{enumitem}
\usepackage{caption}

\usepackage{xcolor,colortbl}
\definecolor{grey}{rgb}{0.89, 0.898, 0.906}

\usepackage{amsmath}
\usepackage{url}
\usepackage{hyperref}

\usepackage{natbib}

\renewcommand\cite{\citep}	

\input{includes/variables.tex}

\begin{document}

\title{{\DefenseShortFullName}}


\author{
  Md Athikul Islam, Edoardo Serra \\
  Boise State University, Boise, Idaho, USA \\
  \texttt{\{mdathikulislam,edoardoserra\}@boisestate.edu}
  \and
  Sushil Jajodia \\
  George Mason University, Fairfax, Virginia, USA \\
  \texttt{jajodia@gmu.edu}
}




\maketitle


\begin{abstract}
Adversarial attacks pose significant challenges to deep neural networks (DNNs) such as Transformer models in natural language processing (NLP). This paper introduces a novel defense strategy, called \textbf{\DefenseShortName}, which enhances adversarial robustness by learning and reasoning on the training classification distribution. {\DefenseShortName} identifies potentially malicious instances deviating from the distribution, transforms them into semantically equivalent instances aligned with the training data, and employs ensemble techniques for a unified and robust response. By conducting extensive experiments, we show that {\DefenseShortName} outperforms state-of-the-art defenses in \emph{accuracy under attack} and \emph{attack success rate} metrics. Additionally, it requires a high number of queries per attack, making the attack more challenging in real scenarios. The ablation study shows that our approach integrates transfer learning, a generative/evolutive procedure, and an ensemble method, providing an effective defense against NLP adversarial attacks. 


\end{abstract}





\section{Introduction}
\label{sec:introduction}
\input{includes/introduction}

\section{Related Work}
\label{sec:related_work}
\input{includes/related_work}

\section{Methodology}
\label{sec:methodology}
\input{includes/methodology}

\section{Experiments}
\label{sec:experiments}
\input{includes/experiments}

\section{Conclusion}
\label{sec:conclusion}
\input{includes/conclusion}

\newpage
\section{Limitations}
Our defense strategy, \textbf{\DefenseShortName}, shows efficacy in mitigating a range of NLP attacks on {\RoBertaModelName} and {\BertModelName} models. However, further research is essential to address the following limitations:

\begin{itemize}
\item \textbf{\DefenseShortName} operates under the assumption that successful attacks produce instances lying outside the distribution of the training set, where the victim model lacks explicit training. Our experiments confirm the validity of this assumption, enabling \textbf{\DefenseShortName} to advance the current state of defensive modeling. However, further studies are needed to identify attacks more aligned with the training classification distribution of the victim model. This can potentially lead to new attack models.
\item While \textbf{\DefenseShortName} relies on learning the training classification distribution, we employ a Gaussian Mixture Model to process text embedding representations. Exploring alternative or more sophisticated approaches could enhance performance. Experimenting with different text embedding models and anomaly detection techniques such as autoencoders could lead to better ways to learn such distribution.
\item Our defense model's effectiveness relies upon the quality of paraphrasing. Investigating additional paraphrasing methodologies could improve its performance.
\item In specialized NLP tasks involving scientific terminology, such as automatic scientific claim verification, the presence of synonymous words can compromise the statement's meaning. Consequently, our approach, like others, necessitates further investigation and refinement in such contexts.
\end{itemize}


\bibliographystyle{acl_natbib}
\bibliography{biblio}


\end{document}

%% file: includes/variables.tex
\newcommand\DefenseShortName{GenFighter}
\newcommand\DefenseFullName{A Generative and Evolutive Textual Attack Removal}

\newcommand\DefenseShortFullName{{\DefenseShortName}: {\DefenseFullName}}

\newcommand\ImdbDataset{IMDB}
\newcommand\AgnewsDataset{AG’s News}
\newcommand\SstDataset{SST-2}

\newcommand\RoBertaModelName{RoBERTa}
\newcommand\BertModelName{BERT}

\newcommand\CleanAccuracyMetricName{Clean\%}
\newcommand\AuaMetricName{Aua\%}
\newcommand\AttackSuccessMetricName{Suc\%}
\newcommand\QueryCountMetricName{\#Query}

\newcommand\NumOfPara{N_p}
\newcommand\Threshold{\tau}
\newcommand\Round{R}
\newcommand\NumOfComponents{N_c}
\newcommand\NumOfCandidates{K}
\newcommand\AlphaPercentile{\alpha}

\newcommand\Paraphraser{p}
\newcommand\MixerModel{m}
\newcommand\TrainData{X}
\newcommand\InputText{x}

\newcommand\AnomalyScores{S}
\newcommand\Paraphrases{P}
\newcommand\TraininDataFeatures{E}
\newcommand\TopKParaphrases{P_K}
\newcommand\TopKAnomalyScores{S_K}  
\newcommand\AnomalyScoresWeights{W_K}
\newcommand\TopKConfidences{C_K}

\newcommand\PwwsAttack{PWWS}
\newcommand\TfAttack{TextFooler}
\newcommand\BaAttack{BERT-Attack}

%% file: includes/introduction.tex
The advent of deep neural networks (DNNs) has achieved significant success in natural language processing (NLP) tasks, nevertheless, they are prone to adversarial attacks designed to exploit their inherent vulnerabilities \cite{alzantot-etal-2018-generating, jin2020textfooler, zeng-etal-2021-openattack}. These attacks involve manipulating the input text with the goal of preserving its semantic meaning while deceiving the target model into generating incorrect outputs. In the context of NLP classification models, such attacks alter the input text to a semantically equivalent version, thereby influencing the model's classification output to differ from the original input text's classification.

Adversarial attacks often involve manipulations at the character \cite{gao_deepwordbug_2018, li2018textbugger}, word \cite{ebrahimi-etal-2018-hotflip, ren2019generating}, or sentence  \cite{ribeiro-etal-2018-semantically, maheshwary-etal-2021-strong} level. As already pointed out by \cite{ wang-etal-2023-rmlm}, character-level and sentence-level attacks are less effective than word-substitution attacks. 
Therefore, we consider only word-substitution attacks.


To defend against word-substitution adversarial attacks, an intuitive approach is to generate adversarial examples through synonym substitutions and training the target model with perturbed instances \cite{ren2019generating, jin2020textfooler, li-etal-2020-bert-attack}. The impracticality of these methods arises from the exponential increase in the number of potential perturbations. Furthermore, more robust adversarial attacks, deviating from these adversarial training strategies, pose significant challenges to these training-based defense methods \cite{si-etal-2021-better}. Some recent works utilize regularization techniques \cite{10.5555/3524938.3525366, liu-etal-2022-flooding}, or ensemble or randomization methods \cite{zeng2021certified, wang-etal-2023-rmlm} to improve the model's robustness.
However, their experimental results in terms of accuracy under attack and attack success rate metrics show room for improvement.  

In this work, we propose an innovative approach, called \textbf{\DefenseShortName} (see Figure \ref{fig:possibleWordModelSemantics}), to extensively enhance adversarial robustness against strong textual attack strategies. Our method operates under the assumption that a successful adversarial attack occurs when the resulting instance lies outside the established training distribution. 
While it is possible that attacks aligned with the training distribution exist, our experiments demonstrate that this assumption leads to improvement beyond the current state-of-the-art. 
Consequently, we design {\DefenseShortName} to learn the training data distribution and identify instances that deviate from this distribution as potentially malicious.
It then searches, through an evolutive process, for semantically equivalent instances that better align with the established distribution. Ultimately, {\DefenseShortName} ensembles the classifications of all generated semantically equivalent instances to provide a robust and unified answer.

In summary, our contributions are as follows:

\begin{itemize}
  \item We design {\DefenseShortName}, a novel defense strategy that tackles adversarial attacks in NLP tasks. The distinctive feature of {\DefenseShortName} lies in its ability to learn and reason on the training classification distribution. It transforms an anomalous instance with respect to the training classification distribution into several semantically equivalent ones more aligned with the training distribution and combines their classification results for a uniquely robust classification.
  
  \item Our experiments show that {\DefenseShortName} often exhibits superior accuracy under attack and a lower attack success rate with respect to the state-of-the-art defensive models for {\RoBertaModelName} and {\BertModelName} as victim models, and the three most popular attacks: {\PwwsAttack}, {\TfAttack}, and {\BaAttack}. Additionally, even the number of queries required to conduct the attack is among the largest compared to the state-of-the-art, implying a further degree of difficulty in conducting attacks on our defensive model. We also demonstrate that our approach outperforms the state-of-the-art in terms of the transferability of attacks, i.e., when attacks are generated against different victim models.
  
  \item The ablation study shows that each sub-component of \DefenseShortName\ is crucial to achieving its high performance. Our approach represents an effective synergistic integration of transfer learning, generative/evolutive procedures, and ensemble modes in the context of a defensive strategy for NLP adversarial attacks.
  

\end{itemize}

%% file: includes/related_work.tex
In NLP tasks, modern adversarial attack methods employ word substitution strategies to generate adversarial examples \cite{garg-2020-bae, ren2019generating, jin2020textfooler, li-etal-2020-bert-attack}. These attack models identify vulnerable words by analyzing the logit output of the victim model and substitute words using synonyms, similar embedding vectors, or transformer models.

Some defense methods are designed based on insights gained from understanding attack methods and others work independent of attack strategies. Empirical defense methods such as adversarial data augmentation (ADA)  \cite{ren2019generating, jin2020textfooler} augment training data with adversaries using their respective attack methods to train the victim model. In addition, gradient-based adversarial training is another type of empirical defense method that involves adding adversarial perturbations to the input's word embeddings during the training phase \cite{Zhu2020FreeLB, li-etal-2021-searching}. Unfortunately, empirical defense methods are highly vulnerable to robust adversarial attacks \cite{li-etal-2020-bert-attack}. In contrast, certified defense methods \cite{huang-etal-2019-achieving, jia-etal-2019-certified} ensure robustness by establishing a certified space within the range of synonyms employed by adversaries. An example of a certified defense is the generation of the convex hull spanned by the word embeddings of a word and its synonyms \cite{zhou2021defense}. A major drawback of certified defense methods is that defenders must know beforehand how an attacker substitutes synonyms.

Randomized masking of words, combined with voting ensemble strategies operates without requiring knowledge of how adversaries generate synonyms. Unfortunately, these methods exhibit unsatisfactory results on larger datasets \cite{zeng2021certified, li-etal-2021-searching}. Additionally, a regularization technique preventing further reduction of the training loss only works on defending overfitted models trained on smaller datasets \cite{liu-etal-2022-flooding}. 

The robustness of {\DefenseShortName} stems from its ability to work on the training classification distribution, coupled with its adept use of randomization and ensemble strategies, allowing it to operate on any datasets independently of the attacker's strategies.

\begin{figure*}[t!]
\centerline{\includegraphics[width=1\textwidth]{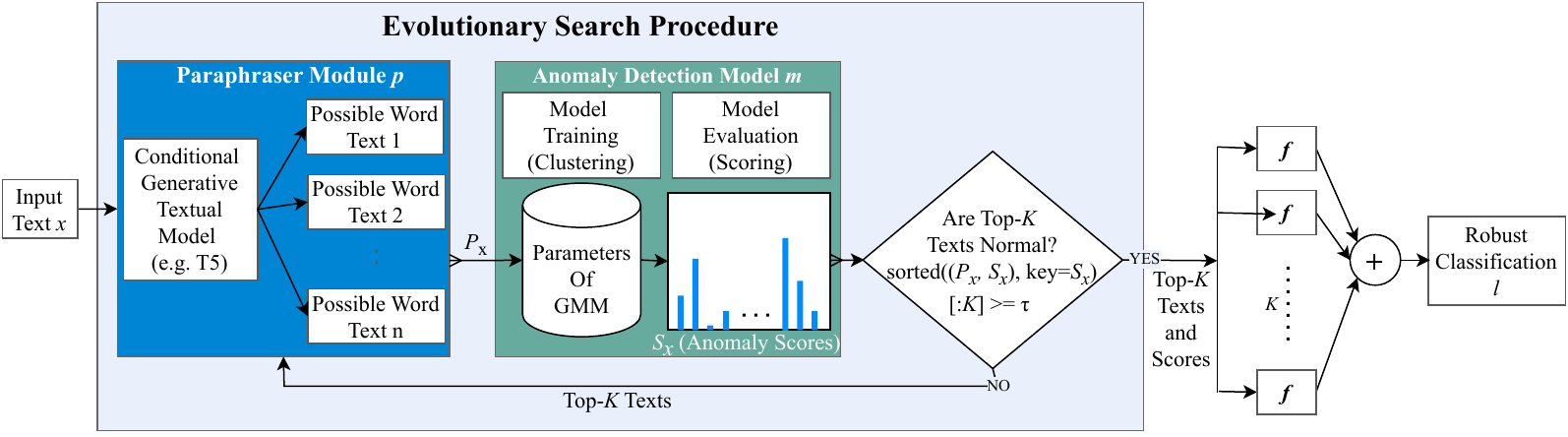}}
\caption{An overview of {\DefenseShortName}. The paraphraser (conditional generative textual model) module processes input texts and forwards input texts and their paraphrases $\Paraphrases_x$ to the anomaly detection model. The anomaly detection model assigns normality scores $\AnomalyScores_x$ and selects top-$\NumOfCandidates$ normal texts. These selected candidate texts are either looped back into the paraphraser module or directed to the target model contingent on meeting the threshold $\Threshold$. Finally, the victim model performs a weighted mean prediction on the top-$\NumOfCandidates$ normal texts and scores.}
    \label{fig:possibleWordModelSemantics}
\end{figure*}

%% file: includes/methodology.tex
We propose the method \textbf{{\DefenseShortName}} that given a target model $f$, the model to defend, it generatively and evolutively removes attacks from text in NLP classification tasks. As described in Figure~\ref{fig:possibleWordModelSemantics}, {\DefenseShortName} employs a paraphraser module for text generation along with an anomaly detection module to characterize the distribution of the training data. Subsequently, our method integrates these modules with an evolutionary search process, that generates paraphrased (perturbed) semantically equivalent text candidates until they align with the training classification distribution. Finally, during the inference phase, an ensemble weighted mean classification is executed on the candidate texts using the target model $f$.  Algorithm \ref{alg:prediction_procedure} provides all the details of our approach, and contains three functions: (1) training, (2) evolutionary search, and (3) inference. A detailed description of each function is provided in the following sections.

\begin{algorithm*}[!t]
\small
\caption{{Training, evolution and inference procedure of  \DefenseShortName}}\label{alg:prediction_procedure}

\resizebox{1\linewidth}{!}{
\begin{minipage}{\linewidth}

\begin{algorithmic}[1]

\Statex \textbf{Initialize:} number of paraphrases $\NumOfPara$, number of best candidates $\NumOfCandidates$, evolution iteration $\Round$

\Statex

\Function{$\mathbf{training}$}{$f, \Paraphraser, \MixerModel, \TrainData$} 
\Statex \textbf{Input:} pre-trained target model $f$, paraphraser model $\Paraphraser$, GMM model $\MixerModel$, training instances $\TrainData$
\Statex \textbf{Output:} fine-tuned target model $f$, trained GMM model $\MixerModel$, threshold $\Threshold$ 
\State {Generate paraphrases} $\Paraphrases \leftarrow \Paraphraser(\TrainData, \NumOfPara)$
\State $\TrainData \leftarrow \TrainData \cup \Paraphrases$
\State {Fine-tune $f$ on $\TrainData$}
\State {Compute features $\TraininDataFeatures$ for each instance of $\TrainData$ from model $f$} 
\State {Train GMM model ${\MixerModel}$ on features} $\TraininDataFeatures$
\State {Compute log-likelihood normal scores} $\AnomalyScores \leftarrow \MixerModel(\TraininDataFeatures)$
\State {Calculate threshold} $\Threshold$ {as $\AlphaPercentile$-percentile of} $\AnomalyScores$ 
\State \Return {(}$f, {\MixerModel}, \Threshold${)}
\EndFunction

\Statex

\Function{$\mathbf{evolution}$}{$f, \Paraphraser, \MixerModel, \Threshold, \InputText$} 
\Statex \textbf{Input:} defined in inference function
\Statex \textbf{Output:} top-$\NumOfCandidates$ texts and scores
\ForAll{$round=1\dots \Round$}
\State {Generate paraphrases} ${\Paraphrases}_x \leftarrow \Paraphraser(\InputText, \NumOfPara)$
\State ${\Paraphrases}_x \leftarrow {\Paraphrases}_x \cup x$
\State {Compute features ${\TraininDataFeatures}_x$ for each instance of ${\Paraphrases}_x$ from model $f$} 
\State {Normal scores} ${\AnomalyScores}_x  \leftarrow \MixerModel({\TraininDataFeatures}_x)$
\State $\TopKParaphrases, \TopKAnomalyScores \leftarrow sorted(({\Paraphrases}_x, {\AnomalyScores}_x), key={\AnomalyScores}_x)[: \NumOfCandidates]$
\If{$\TopKAnomalyScores \geq \Threshold$}
    \State \textbf{break}
\EndIf
\State $x \leftarrow \TopKParaphrases$
\EndFor


\State \Return {(}${\TopKParaphrases}, \TopKAnomalyScores${)}
\EndFunction

\Statex

\Function{$\mathbf{inference}$}{$f, \Paraphraser, \MixerModel, \Threshold, \InputText$} 
\Statex \textbf{Input:} fine-tuned target model $f$, paraphraser model $\Paraphraser$, trained GMM model $\MixerModel$, threshold $\Threshold$, input text $\InputText$
\Statex \textbf{Output:} classification label $l$
\State $\TopKParaphrases, \TopKAnomalyScores \leftarrow evolution(f, \Paraphraser, \MixerModel, \Threshold, \InputText)$ 

\State {Weights} $W_K \leftarrow \frac{ \frac{S_K - S_{K_{\text{min}}}}{S_{K_{\text{max}}} - S_{K_{\text{min}}}} } { \sum_{i=1}^{K} \frac{S_i - S_{K_{\text{min}}}}{S_{K_{\text{max}}} - S_{K_{\text{min}}}} }$

\State {Confidences} $C_\NumOfCandidates \leftarrow f(\TopKParaphrases)$
\State $l \leftarrow {argmax}(W_\NumOfCandidates.dot(C_\NumOfCandidates))$
\State \Return $l$
\EndFunction

\end{algorithmic}

\end{minipage}
}

\end{algorithm*}
\subsection{Training}

The training phase includes generating paraphrases, fine-tuning the target model, training the anomaly detection model, and defining threshold selection.
Our defense mechanism first introduces a conditional generative textual model $\Paraphraser$, such as a T5 transformer \cite{raffel2019exploring}, with the task of generating paraphrases from input texts. Utilizing this paraphraser module, {\DefenseShortName} augments the training data $\TrainData$ with their paraphrased instances $\Paraphrases$ to fine-tune the target model (Algorithm \ref{alg:prediction_procedure}, lines 2-4). To enhance the diversity of paraphrases, it employs stochastic sampling, drawing from the model's probability distribution to predict the next output token. This approach ensures the generation of varied paraphrases for the same text across different instances. For each text input, we denote hyperparameter the number of paraphrases $\NumOfPara$. \DefenseShortName\ also uses this paraphraser model during the evolutionary search procedure.
It trains an anomaly detection model $\MixerModel$ specifically the Gaussian Mixture model (GMM) to learn the distribution of the training data (Algorithm \ref{alg:prediction_procedure}, lines 5-6). First, it obtains feature representations $\TraininDataFeatures$ for training instances by extracting information from the final layer of the target model $f$ (which in our experiments will be a transformer). Then, it determines the optimal number of clusters or components $\NumOfComponents$ on the training data to initialize GMM using the Bayesian Information Criterion (BIC) \cite{schwarz1978estimating} principle. Following the initialization of the GMM with $\NumOfComponents$ components, it trains the model $\MixerModel$ (GMM) using the feature representation $\TraininDataFeatures$. At this stage, the model $\MixerModel$ is prepared to compute the log-likelihood scores $\AnomalyScores$ from $\TraininDataFeatures$, indicating the proximity of a given sample to the training data. Higher scores correspond to samples that align more closely with the training classification distribution.
After training the anomaly detection model, the algorithm computes a threshold $\Threshold$ to determine if a sample is aligned with the training classification distribution or is an outlier. The threshold $\Threshold$ is computed as the $\AlphaPercentile$-percentile of scores $\AnomalyScores$. Samples with scores below this threshold are identified as outliers within the training distribution. $\AlphaPercentile$ is a hyperparameter of the function training. 

\subsection{Evolutionary Search}

The evolutionary search procedure perturbs outlier examples, by gradually aligning them with the training classification distribution during the inference phase. This procedure uses the pre-trained paraphraser module ${\Paraphraser}$ for text generation and utilizes the trained anomaly model $\MixerModel$ for normality scoring. 

Let $\Round$ be the number of iterations and $\NumOfCandidates$ be the number of best candidate texts. During each iteration, the paraphraser module ${\Paraphraser}$ generates semantically equivalent texts $\Paraphrases_x$ for input text $\InputText$ (Algorithm \ref{alg:prediction_procedure}, line 13). It extends $\Paraphrases_x$ by $\InputText$ for top-$\NumOfCandidates$ candidate texts selection and generates feature representations $\TraininDataFeatures_x$ from $f$. Subsequently, it computes the normal likelihood scores $\AnomalyScores_x$ for the $\Paraphrases_x$ using $\MixerModel$. Finally, it sorts these scores $\AnomalyScores_x$ alongside their corresponding texts $\Paraphrases_x$ in descending order based on the scores $\AnomalyScores_x$. Slicing the sorted result by $\NumOfCandidates$ produces top-$\NumOfCandidates$ candidate texts $\TopKParaphrases$ and their scores $\TopKAnomalyScores$ (Algorithm \ref{alg:prediction_procedure}, line 17). The iteration terminates when scores $\TopKAnomalyScores$ surpass the threshold $\Threshold$ or when the maximum iteration $\Round$ is reached.


\subsection{Inference}

The inference function integrates the evolutionary search function and an ensemble-weighted mean classification procedure for robust text classification. During inference, it initially receives the input text and invokes the evolutionary search procedure, gathering the top-$\NumOfCandidates$ candidate texts $\TopKParaphrases$ along with their scores $\TopKAnomalyScores$. Subsequently, it applies min-max normalization to the $\TopKAnomalyScores$, yielding normalized weights $\AnomalyScoresWeights$ (Algorithm \ref{alg:prediction_procedure}, line 27).
To emphasize the relative importance of each score, it normalizes the weights within the range [0, 1] by dividing them by their sum. 

Then, the target model $f$ generates confidences $\TopKConfidences$ for the top-$\NumOfCandidates$ candidate texts $\TopKParaphrases$. Consequently, it performs a dot product on $\AnomalyScoresWeights$ and $\TopKConfidences$ to derive the final confidence of the input text $\InputText$. Applying the $argmax$ function to the final confidence produces the final robust classification $l$.

%% file: includes/experiments.tex
\begin{table*}[!h]
\centering
\small
\begin{tabular}{ccccc}
\hline
\hline
\textbf{Dataset} & \textbf{Training Set} & \textbf{Test Set} & \textbf{\# of Words} & \textbf{\# of Classes}
\\
\hline
{\ImdbDataset} & 25,000 & 25,000 & 268 & 2 \\
{\AgnewsDataset} & 120,000 & 7,600 & 40 & 4 \\
{\SstDataset} & 6,920 & 1821 & 19 & 2 \\
\hline
\hline
\end{tabular}

%

\caption{
Statistics of Datasets.
}
\label{table:dataset-statistics}
\end{table*}
In this section, we show our experimental evaluation of \DefenseShortName\ in comparison to state-of-the-art defense methods on three different datasets and against three different adversarial attacks. More specifically in the following section, we provide details about our experiments regarding datasets, attack methods, victim models, and state-of-the-art defense methods considered. Then we provide the implementation settings and a discussion of the main results. In addition, we present an ablation study showing the importance of each sub-component of {\DefenseShortName}. Moreover, we provide results of robustness to attack transferability. 

\subsection{Datasets}

We conduct extensive analysis on three benchmark classification datasets:  \textbf{\ImdbDataset} \cite{maas-etal-2011-learning}, \textbf{\AgnewsDataset} \cite{NIPS2015_250cf8b5}, and
\textbf{\SstDataset} \cite{socher-etal-2013-recursive}. {\ImdbDataset} is a document-level movie review dataset for sentiment analysis. {\AgnewsDataset}  is a sentence-level dataset for multi-class news classification, having four types of classes: World, Sports, Business, and Science. {\SstDataset} is a binary classification dataset at the sentence level for sentiment analysis. The statistics of these three datasets including \# of examples in training and test sets, \# of classes, and the average \# of words are listed in Table \ref{table:dataset-statistics}.

\begin{table*}[ht!]
\centering
\resizebox{\textwidth}{!}{\begin{tabular}{l|l|l|c|ccc|ccc|ccc}

\hline
\hline
\multirow{2}{*}{Datasets} & \multirow{2}{*}{DNNs} & \multirow{2}{*}{Methods} & \multirow{2}{*}{\CleanAccuracyMetricName} & 
\multicolumn{3}{c|}{\PwwsAttack}& 
\multicolumn{3}{c|}{\TfAttack}& 
\multicolumn{3}{c}{\BaAttack}

\\
& & & & \AuaMetricName & \AttackSuccessMetricName & \QueryCountMetricName & \AuaMetricName & \AttackSuccessMetricName & \QueryCountMetricName & \AuaMetricName & \AttackSuccessMetricName  & \QueryCountMetricName
\\

\hline

\multirow{12}{*}{\ImdbDataset} & \multirow{6}{*}{\RoBertaModelName} &Original&93.82&14.4&84.6&1529&8.6&90.8&554&37.8&59.6&1623
\\
&&RanMASK \cite{zeng2021certified}&93.37&45.5&51.3&1566&45.3&51.6&859&42.3&54.8&1787
\\
&&FreeLB++ \cite{li-etal-2021-searching}&\textbf{94.32}&35.7&62.3&1575&19.0&79.9&681&\textbf{45.2}&\textbf{52.3}&1723
\\
&&Flooding-X \cite{liu-etal-2022-flooding}&93.90&45.5&51.2&1575&22.0&76.4&735&44.2&52.6&\textbf{1904}
\\
&&RMLM \cite{wang-etal-2023-rmlm}&92.77&37.3&59.5&1569&47.4&48.6&923&22.2&75.9&1209
\\
&&\textbf{\DefenseShortName}&92.37&\textbf{64.8}&\textbf{30.2}&\textbf{1606}&\textbf{69.2}&\textbf{25.4}&\textbf{1023}&43.0&53.7&1652
\\

\cline{2-13}

& \multirow{6}{*}{\BertModelName} &Original&92.05&22.1&75.9&1544&8.8&90.4&604&15.3&83.3&577
\\
&&RanMASK \cite{zeng2021certified}&91.95&46.9&44.8&1536&48.1&43.7&916&\textbf{51.5}&\textbf{39.4}&\textbf{2069}
\\
&&FreeLB++ \cite{li-etal-2021-searching}&91.29&26.4&70.8&1547&22.1&75.6&715&34.0&62.4&1424
\\
&&Flooding-X \cite{liu-etal-2022-flooding}&\textbf{92.35}&49.0&47.0&1583&31.9&65.5&809&43.5&52.9&2028
\\
&&RMLM \cite{wang-etal-2023-rmlm}&90.80&28.9&67.8&1568&44.0&50.9&949&23.1&74.3&1240
\\
&&\textbf{\DefenseShortName}&91.10&\textbf{58.8}&\textbf{35.5}&\textbf{1588}&\textbf{61.8}&\textbf{32.2}&\textbf{1000}&44.0&51.7&1900
\\
\hline

\multirow{12}{*}{\AgnewsDataset} & \multirow{6}{*}{\RoBertaModelName} & Original&94.57&54.9&42.1&260&49.6&47.7&167&58.3&38.5&552
\\
&&RanMASK \cite{zeng2021certified}&94.75&64.9&31.2&264&64.2&31.9&179&64.0&32.1&585
\\
&&FreeLB++ \cite{li-etal-2021-searching}&33.37&0.4&98.8&242&6.7&79.5&107&0.5&98.5&101
\\
&&Flooding-X \cite{liu-etal-2022-flooding}&\textbf{94.84}&55.7&40.9&262&49.5&47.5&171&58.9&37.5&566
\\
&&RMLM \cite{wang-etal-2023-rmlm}&93.86&67.8&27.3&259&76.3&18.2&189&44.4&52.4&494
\\
&&\textbf{\DefenseShortName}&94.29&\textbf{78.3}&\textbf{17.1}&\textbf{268}&\textbf{79.7}&\textbf{15.7}&\textbf{195}&\textbf{65.8}&\textbf{30.4}&\textbf{605}
\\

\cline{2-13}

& \multirow{6}{*}{\BertModelName} & Original&94.61&57.4&39.1&261&46.7&50.4&168&46.8&50.3&229
\\
&&RanMASK \cite{zeng2021certified}&94.79&63.4&32.6&263&62.2&34.0&178&64.0&32.1&580
\\
&&FreeLB++ \cite{li-etal-2021-searching}&\textbf{95.22}&65.0&31.7&264&58.2&38.9&178&61.9&35.0&566
\\
&&Flooding-X \cite{liu-etal-2022-flooding}&94.59&64.7&31.3&264&56.1&40.5&175&61.9&34.3&569
\\
&&RMLM \cite{wang-etal-2023-rmlm}&93.79&69.3&24.9&259&77.1&\textbf{16.6}&191&41.8&54.7&558
\\
&&\textbf{\DefenseShortName}&94.37&\textbf{74.2}&\textbf{20.7}&\textbf{268}&\textbf{77.5}&17.2&\textbf{194}&\textbf{66.6}&\textbf{28.9}&\textbf{604}
\\
\hline

\multirow{12}{*}{\SstDataset} & \multirow{6}{*}{\RoBertaModelName} & Original&94.12&35.2&62.9&115&33.6&64.6&67&25.9&72.7&138
\\
&&RanMASK \cite{zeng2021certified}&93.30&33.7&63.9&115&41.5&55.6&70&21.5&77.0&125
\\
&&FreeLB++ \cite{li-etal-2021-searching}&51.84&3.1&93.9&107&4.4&91.3&44&1.5&97.0&47
\\
&&Flooding-X \cite{liu-etal-2022-flooding}&\textbf{95.11}&37.3&60.7&115&35.7&62.3&68&19.7&79.2&117
\\
&&RMLM \cite{wang-etal-2023-rmlm}&92.51&45.1&51.2&116&58.7&36.5&80&19.4&78.9&98
\\
&&\textbf{\DefenseShortName}&93.14&\textbf{51.2}&\textbf{45.5}&\textbf{117}&\textbf{60.9}&\textbf{35.1}&\textbf{81}&\textbf{27.6}&\textbf{70.6}&\textbf{155}
\\

\cline{2-13}

& \multirow{6}{*}{\BertModelName} & Original&91.71&28.1&69.5&113&26.3&71.4&61&14.3&84.5&62
\\
&&RanMASK \cite{zeng2021certified}&91.32&29.0&68.3&114&35.2&61.3&67&20.3&77.8&122
\\
&&FreeLB++ \cite{li-etal-2021-searching}&\textbf{92.59}&40.3&56.4&116&37.5&59.5&68&\textbf{32.0}&\textbf{65.4}&148
\\
&&Flooding-X \cite{liu-etal-2022-flooding}&91.49&31.0&66.2&114&32.0&65.1&64&20.4&77.7&113
\\
&&RMLM \cite{wang-etal-2023-rmlm}&85.94&36.9&57.6&116&52.1&40.1&80&14.2&83.6&100
\\
&&\textbf{\DefenseShortName}&91.22&\textbf{47.4}&\textbf{47.2}&\textbf{117}&\textbf{56.1}&\textbf{37.5}&\textbf{81}&25.8&71.3&\textbf{157}
\\

\hline
\hline
\end{tabular}}
\caption{Experimental results of {\DefenseShortName} in comparison of state-of-the-art methods on three datasets against three word-substitution attacks where all models are trained on \underline{\RoBertaModelName} and \underline{\BertModelName}. The best
performance is marked in \textbf{bold}.}
\label{table:experimental-results}
\end{table*} 

\begin{table*}[ht!]
\centering
\resizebox{\textwidth}{!}{\begin{tabular}{l|l|c|ccc|ccc|ccc}
\hline
\hline
\multirow{2}{*}{Datasets} & \multirow{2}{*}{Methods} & \multirow{2}{*}{\CleanAccuracyMetricName} & 
\multicolumn{3}{c|}{\PwwsAttack}& 
\multicolumn{3}{c|}{\TfAttack}& 
\multicolumn{3}{c}{\BaAttack}
\\
& & & \AuaMetricName & \AttackSuccessMetricName & \QueryCountMetricName & \AuaMetricName & \AttackSuccessMetricName & \QueryCountMetricName & \AuaMetricName & \AttackSuccessMetricName  & \QueryCountMetricName
\\

\hline

\multirow{4}{*}{\AgnewsDataset}&Original&\textbf{94.57}&54.9&42.1&260&49.6&47.7&167&58.3&38.5&552
\\
&T5 Paraphraser&94.55&48.4&48.4&259&45.4&51.6&168&54.9&41.5&545
\\
&Original + T5 Paraphraser&94.55&67.1&29.0&264&59.7&36.8&179&61.8&34.6&594
\\
&\textbf{\DefenseShortName}&94.29&\textbf{78.3}&\textbf{17.1}&\textbf{268}&\textbf{79.7}&\textbf{15.7}&\textbf{195}&\textbf{65.8}&\textbf{30.4}&\textbf{605}
\\

\hline

\multirow{4}{*}{\SstDataset}&Original&94.12&35.2&62.9&115&33.6&64.6&67&25.9&72.7&138
\\
&T5 Paraphraser&92.86&32.7&64.6&115&31.1&66.3&64&24.9&73.1&134
\\
&Original + T5 Paraphraser&\textbf{94.67}&38.5&59.6&115&35.8&62.5&66&27.2&71.5&144
\\
&\textbf{\DefenseShortName}&93.14&\textbf{51.2}&\textbf{45.5}&\textbf{117}&\textbf{60.9}&\textbf{35.1}&\textbf{81}&\textbf{27.6}&\textbf{70.6}&\textbf{155}
\\

\hline
\hline
\end{tabular}}
\caption{ Ablation study findings evaluated using three word-substitution attacks on the {\AgnewsDataset} and {\SstDataset} datasets. The best
performance is marked in \textbf{bold}.}
\label{table:ablation-study}
\end{table*} 

\subsection{Attack Methods}

We employ the same three strong attacks used in \cite{wang-etal-2023-rmlm} to evaluate the efficiency of \DefenseShortName\ against all baseline attacks. Such selection of attack strategies is based on their proficiency in preserving semantics and showcasing superior attack effectiveness.

{\PwwsAttack} \cite{ren2019generating} uses WordNet\footnote{\url{https://wordnet.princeton.edu/}} to construct synonyms as substitution words and leverages the change in the model's probability and word saliency for the word replacement order. \citet{jin2020textfooler} propose {\TfAttack} attack which determines the importance score of a word by accumulating probability changes before and after deleting the word. The attack further uses synonym replacement through a set of cosine-similar embedding vectors. {\BaAttack} \cite{li-etal-2020-bert-attack} utilizes the change of logit output of the victim model for finding vulnerable words and generates semantic-preserving word replacement using {\BertModelName}. 


\subsection{Victim Models and State-of-the-art Defense Methods}

We perform experiments on two types of DNNs as victim models: {\RoBertaModelName}-base \cite{liu2020roberta} and BERT-base-uncased \cite{devlin-etal-2019-bert}. 

For the state-of-the-art defense comparison, we compare \DefenseShortName\ with one randomized smoothing model, one gradient-guided adversarial training model, one overfitting-based model, and one randomized masked language model.

\begin{description}[style=unboxed, leftmargin=0cm]

   \item[RanMASK] \cite{zeng2021certified} introduces a new randomized smoothing technique that masks input texts repeatedly and performs prediction with a "majority vote".

   \item[FreeLB++] \cite{li-etal-2021-searching} a gradient-based training method, enhances FreeLB \cite{Zhu2020FreeLB} by expanding the search steps, resulting in a larger $l2$-norm search region.

   \item[Flooding-X] \cite{liu-etal-2022-flooding} restricts any further reduction in the training loss and improves the model's generalization by introducing a flooding criterion.

   \item[RMLM] \cite{wang-etal-2023-rmlm} tries to defend an attack by randomly introducing noise to sentences and then correcting the contexts of these corrupted perturbations. At last, it filters out adversarial samples.

\end{description}

\subsection{Evaluation Metrics}
We report four evaluation metrics to assess the robustness of the aforementioned defense methods against different adversarial attacks.

\begin{description}[style=unboxed, leftmargin=0cm]
   \item[Clean Accuracy (\CleanAccuracyMetricName)] is the model's classification accuracy on the original full clean test dataset.

   \item[Accuracy Under Attack (\AuaMetricName)] is the prediction accuracy under an adversarial attack. A higher {\AuaMetricName} reflects the better performance of a defender.

   \item[Attack Success Rate (\AttackSuccessMetricName)] is the ratio of \# of successfully pertured texts to the \# of total attempted texts. A lower {\AttackSuccessMetricName} suggests a model is more robust.

   \item[Number of Queries (\QueryCountMetricName)] is the average number of queries the attacker attempts for each attack. A defending model should produce higher {\QueryCountMetricName} in order not to be easily compromised.
\end{description}

\subsection{Implementation Settings}

We leverage OpenAttack \cite{zeng-etal-2021-openattack} to perform all attacks and compare \DefenseShortName\ against state-of-the-art defense methods. All defense methods are reproduced using open-source code and their pre-defined hyperparameters. We conduct model training on NVIDIA GeForce RTX 4090 and NVIDIA RTX 6000 GPUs. Moreover, we utilize a T5 paraphraser model previously trained with Google PAWS \cite{paws2019naacl}, accessible via the HuggingFace repository\footnote{\url{https://huggingface.co/Vamsi/T5_Paraphrase_Paws}}. To evaluate metrics \AuaMetricName, \AttackSuccessMetricName, and \QueryCountMetricName\ we randomly choose 1,000 samples from the whole test set (similarly to what was done in \cite{wang-etal-2023-rmlm} and others) and we use such set for our approach and the state-of-the-art methods. While we compute clean accuracy (\CleanAccuracyMetricName) on the entire test set.
\subsection{Main Results}

We summarize the performance of {\DefenseShortName} compared to the state-of-the-art defenses in Table \ref{table:experimental-results}. Overall, our proposed approach enhances the robustness of both the {\RoBertaModelName} and {\BertModelName} across all three datasets against three strong adversarial attack strategies.

Our model outperforms the state-of-the-art defense in 15 out of 18 cases (83\%), as measured by {\AuaMetricName}. Against {\PwwsAttack} and {\TfAttack} attacks, our model showcases superior resilience, consistently outmatching all state-of-the-art defense methods in terms of {\AuaMetricName}, {\AttackSuccessMetricName}, and {\QueryCountMetricName}. These metrics are crucial for evaluating the robustness of defense algorithms. Moreover, when subjected to {\BaAttack}, our defense surpasses the state-of-the-art defenses in 3 out of 6 cases. Across all datasets and attacks, the absolute average improvements for {\AuaMetricName}, {\AttackSuccessMetricName}, and {\QueryCountMetricName} are +41.6\%, +37.0\%, and +7.8\%, respectively. Notably, our model often consistently achieves the highest {\QueryCountMetricName}, making it particularly challenging to be compromised. In addition, our defense exhibits an extremely competitive {\CleanAccuracyMetricName} and the difference is marginal. It is important to highlight that FreeLB++ demonstrates significant performance degradation, especially apparent with the {\RoBertaModelName}, on both the {\AgnewsDataset} and {\SstDataset} datasets. As a result, we've chosen to omit comparative evaluations of FreeLB++ on these datasets for the {\RoBertaModelName} model. When considering the most recent state-of-the-art defense methods, namely Flooding-X and RMLM, our model consistently demonstrates superior robustness compared to these methods.

\subsection{Ablation Study}

Table \ref{table:ablation-study} illustrates the individual contributions of each component within {\DefenseShortName} when countering an adversarial attack. The "Original" method refers to the victim DNN model without any defense mechanism. We evaluate the efficacy of our paraphrase module "T5 Paraphraser" as it is used as a standalone victim model with no defense. Additionally, we assess the impact of the paraphraser module within the context of {\DefenseShortName} by "Original + T5 Paraphraser". Finally, we examine the overall performance of the {\DefenseShortName} method. 

We summarize the following findings:
(1) The "T5-Paraphraser" used alone as a victim model does not provide sufficient defense against adversarial attacks, showing a performance decline compared to the "Original" victim method. (2) Combining "T5-Paraphraser" with the "Original" target method enhances overall performance. The absolute average improvements introduced by our "T5-Paraphraser" to the "Original" method are +12.7\%, +10.5\%, and +4.8\% for {\AuaMetricName}, {\AttackSuccessMetricName} and {\QueryCountMetricName} respectively. Furthermore, the enhancement achieved by {\DefenseShortName} over "Original + T5 Paraphraser" is attributed to our anomaly detection model and evolutionary search procedure. In this case, the absolute average gains are +25.3\%, +27.1\%, and +4.4\% for {\AuaMetricName}, {\AttackSuccessMetricName} and {\QueryCountMetricName} respectively.
This shows that all the sub-components of the {\DefenseShortName} method are well integrated and needed to achieve its performance.

\begin{table*}[ht!]
\centering
\small
\begin{tabular}{l|l|c|c|c|c}
\hline
\hline
Datasets & Attack Methods & Original & Flooding-X & RMLM & \textbf{\DefenseShortName}
\\

\hline

\multirow{3}{*}{\AgnewsDataset}&{\PwwsAttack}&54.9*&61.4&80.7&\textbf{84.2}
\\
&{\TfAttack}&49.6*&66.8&83.8&\textbf{87.4}
\\
&{\BaAttack}&58.3*&63.8&70.7&\textbf{75.9}
\\

\hline

\multirow{3}{*}{\SstDataset}&{\PwwsAttack}&35.2*&40.5&53.6&\textbf{56.3}
\\
&{\TfAttack}&33.6*&53.7&55.8&\textbf{61.3}
\\
&{\BaAttack}&25.9*&34.6&48.7&\textbf{50.4}
\\

\hline
\hline
\end{tabular}
\caption{
Transferability assessment on {\AgnewsDataset} and {\SstDataset} datasets against three word-substitution attacks, with all models trained on \underline{\RoBertaModelName}. Asterisk (*) denotes adversarial examples generated using the original model, validated for transferability across other models based on the classification accuracy (\%). The best
performance is marked in \textbf{bold}.}
\label{table:transferability}
\end{table*}

\subsection{Hyperparameter Analysis}

\begin{figure}[h]

    \centering
    \includegraphics[width=0.8\columnwidth]{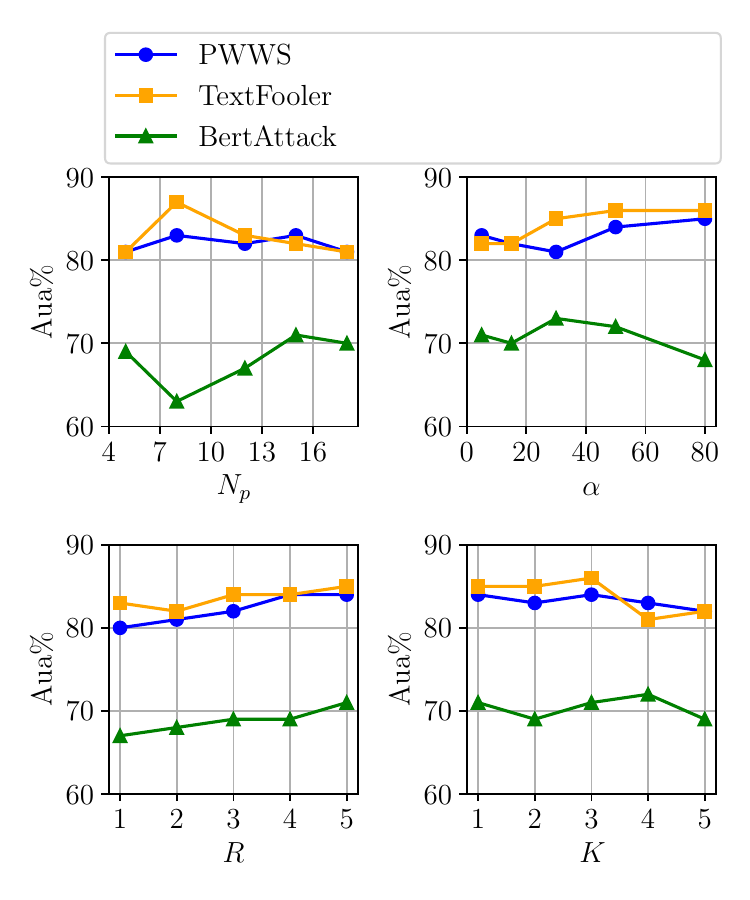}
    \caption{Hyperparameter analysis of {\DefenseShortName} against three word-substitution attacks on the {\AgnewsDataset} dataset. We randomly sample 100 test examples and compute the {\AuaMetricName} for each attack across all hyperparameters.}
    \label{fig:hyperparameter-analysis}
\end{figure}

Figure \ref{fig:hyperparameter-analysis} depicts how various hyperparameters, including the number of paraphrases (${\NumOfPara}$), $\AlphaPercentile$-percentile, number of rounds (${\Round}$), and number of best candidates (${\NumOfCandidates}$), affect the outcome of {\DefenseShortName} in terms of {\AuaMetricName}.

\begin{description}[style=unboxed, leftmargin=0cm]

   \item[Number of Paraphrases] {\DefenseShortName} consistently defends against {\PwwsAttack} and {\TfAttack} attacks across all values of number of paraphrases $\NumOfPara$, whereas its effectiveness against {\BaAttack} improves with higher values of $\NumOfPara$. Opting for higher values of $\NumOfPara$ comes with increased computational costs. Hence, choosing a good trade-off is important.
   
   \item[$\AlphaPercentile$-percentile] We establish threshold ${\Threshold}$ as the $\AlphaPercentile$-percentile value based on training data normality scores. Given the extensive query generation by {\BaAttack} and the challenge of meeting high ${\Threshold}$ values, aligning selected candidates with the training distribution becomes difficult. Notably, our model begins to improve after the 15th and 30th $\AlphaPercentile$-percentiles against {\TfAttack} and {\PwwsAttack}, respectively.

   \item[Number of Rounds] $\Round$ stands out as the most critical parameter in our evolutionary search procedure. With each iteration, {\DefenseShortName} augments the candidate population and refines selection within the search process. Remarkably, {\AuaMetricName} consistently improves with a higher number of $\Round$. 
   This indicates that the ability of our model to contrast attacks depends on the time allocated for the evolutive search process.
   We maintain a constant ${\NumOfCandidates}$ throughout this evaluation.

   \item[Number of Best Candidates] To assess the impact of number of best candidates $\NumOfCandidates$ on {\AuaMetricName}, we maintain $\Round$=5 while varying ${\NumOfCandidates}$. As ${\NumOfCandidates}$ approaches $\Round$, {\AuaMetricName} decreases across all three attacks. Optimal performance is observed with mid values of $\Round$ for ${\NumOfCandidates}$. 

\end{description}

\subsection{Transferability}
Transferability in adversarial attacks involves crafting successful adversarial examples using one model and exploiting them to deceive other models with minimal effort. Table \ref{table:transferability} illustrates how {\DefenseShortName} effectively mitigates the transferability of adversarial attacks. We evaluate the recent state-of-the-art defense methods Flooding-X and RMLM for comparison in mitigating transferability. Our findings reveal that {\DefenseShortName} consistently defends against transferability by achieving the highest classification accuracy (\%) across all cases.

%% file: includes/conclusion.tex
In this work, we have addressed the problem of defending against word-substitution adversarial attacks in natural language processing (NLP) tasks, particularly those targeting Transformer models.

To combat such attacks, we proposed a novel defense strategy, \textbf{\DefenseShortName}, which learns the distribution of training data to identify and mitigate potentially malicious instances. Through an evolutionary process, \DefenseShortName\ generates semantically equivalent alternatives aligned with the training data distribution, and ensembles the classifications outcome of such alternatives to provide a unified and robust outcome.

Our experimental results demonstrate the efficacy of \DefenseShortName\ in achieving superior accuracy under attack scenarios and reducing the success rate of adversarial attacks compared to state-of-the-art defenses, on different datasets. 
Last but not least, our approach forces the attacker models to perform the largest number of queries during the attack among all the competitors, by making such attacks infeasible in real scenarios. 
